\documentclass[dvipsnames]{article} %
\usepackage{colm2024_conference}
\pdfoutput=1
\usepackage{natbib}
\setcitestyle{authoryear,round,citesep={;},aysep={,},yysep={;}}
\let\cite\citep
\usepackage{booktabs}
\usepackage{graphicx}
\usepackage{enumitem}
\usepackage{wrapfig}
\usepackage{algorithm}
\usepackage{algpseudocode}
\usepackage{microtype}
\usepackage{amsmath}
\usepackage{amsthm}
\usepackage{colortbl}
\usepackage[utf8]{inputenc}
\definecolor{lightgray}{rgb}{0.9,0.9,0.9}
\usepackage{caption}
\usepackage{subcaption}
\usepackage{setspace}
\usepackage{url}
\usepackage{multirow}
\usepackage{colortbl}
\usepackage{tabularx}
\usepackage{blindtext}
\usepackage{pgfplots}
\pgfplotsset{compat=1.18} 
\usepackage{tikz}
\usetikzlibrary{er,positioning,bayesnet}
\usepackage{makecell}
\usepackage{tipa}
\usepackage{siunitx}
\usepackage{nicefrac}
\usepackage{tocloft}
\usepackage{listings}
\usepackage{cleveref}
\usepackage{placeins}
\usepackage{rotating}
\usepackage{silence}
\usepackage{tablefootnote}
\usepackage{threeparttable}
\usepackage[raster,skins]{tcolorbox} %
\usepackage{xltabular}
\usepackage{adjustbox}
\usepackage{xurl}
\usepackage[normalem]{ulem}
\useunder{\uline}{\ul}{}
\usepackage{newtxtext,newtxmath}
\usepackage{CJKutf8}
\usepackage{pifont}
\usepackage{xcolor}

\newtheorem{definition}{Definition}

\usepackage{amsmath,amsfonts,bm}

\def\eqref#1{equation~\ref{#1}}

\def\1{\bm{1}}

\DeclareMathAlphabet{\mathsfit}{\encodingdefault}{\sfdefault}{m}{sl}
\SetMathAlphabet{\mathsfit}{bold}{\encodingdefault}{\sfdefault}{bx}{n}

\newcommand*\justify{%
  \fontdimen2\font=0.4em%
  \fontdimen3\font=0.2em%
  \fontdimen4\font=0.1em%
  \fontdimen7\font=0.1em%
  \hyphenchar\font=`\-%
}

\renewcommand{\texttt}[1]{%
  \begingroup
  \ttfamily
  \begingroup\lccode`~=`/\lowercase{\endgroup\def~}{/\discretionary{}{}{}}%
  \begingroup\lccode`~=`[\lowercase{\endgroup\def~}{[\discretionary{}{}{}}%
  \begingroup\lccode`~=`.\lowercase{\endgroup\def~}{.\discretionary{}{}{}}%
  \catcode`/=\active\catcode`[=\active\catcode`.=\active
  \justify\scantokens{#1\noexpand}%
  \endgroup
}

\title{\includegraphics[height=3.5ex]{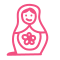}MatryoshkaThinking: Recursive Test-Time Scaling Enables Efficient Reasoning}

\author{
 \centering
 \small{}
Hongwei Chen$^*$  \hspace{4mm} Yishu Lei$^*$  \hspace{4mm} Dan Zhang$^*$  \hspace{4mm} Bo Ke \hspace{4mm} Danxiang Zhu \hspace{4mm} Xuyi Chen \hspace{4mm} Yuxiang Lu \hspace{4mm} Zhengjie Huang \hspace{4mm} Shikun Feng$^{\dag}$ \hspace{4mm} Jingzhou He \hspace{4mm} Yu Sun \hspace{4mm} Hua Wu \hspace{4mm} Haifeng Wang
\\[5pt]
 \centering
 \small{}
  ERNIE Team, Baidu \\[5pt]
   \centering
   \small{}
  \texttt{\{chenhongwei04, leiyishu, zhangdan20, kebo01, zhudanxiang, chenxuyi, luyuxiang, huangzhengjie, fengshikun01, hejingzhou, sunyu02, wu\_hua, wanghaifeng\}@baidu.com}
  \\
  $^{*}$Equal contribution, $^{\dagger}$Corresponding author
}

\makeatletter\def\@abstract{
Test-time scaling has emerged as a promising paradigm in language modeling, wherein additional computational resources are allocated during inference to enhance model performance. Recent approaches, such as DeepConf, have demonstrated the efficacy of this strategy, however, they often incur substantial computational overhead to achieve competitive results. In this work, we propose MatryoshkaThinking, a novel method that significantly reduces computational cost while maintaining state-of-the-art performance. Specifically, MatryoshkaThinking attains a score of 99.79 on AIME25 using only 4\% of the computation required by DeepConf. The core of our approach lies in the recursive exploitation of the model’s intrinsic capabilities in reasoning, verification, and summarization, which collectively enhance the retention of correct solutions and reduce the disparity between Pass@k and Pass@1. Comprehensive evaluations across multiple open-source models and challenging multi-modal reasoning benchmarks validate the effectiveness and generality of our method. These findings offer new insights into the design of efficient and scalable test-time inference strategies for advanced language models.
}\makeatother

\begin{document}
\maketitle

\pagestyle{firstpage}  %
\pagestyle{normalpage}

\section{Introduction}

Large Language Models (LLMs) exhibit remarkable reasoning abilities, which can be further enhanced by methods that adjust behavior during test-time inference. Conventional test-time scaling methods can be broadly classified into two categories: sequential and parallel scaling. Sequential scaling methods, exemplified by SELF-REFINE ~\cite{self-refine}, improve reasoning by iteratively refining responses through self-verification and correction. However, their gains quickly saturate with additional refinement steps, limiting their ability to exploit larger test-time compute budgets. Parallel decoding ~\cite{self-consist} entails generating multiple candidate solutions and selecting the most promising one through strategies such as majority voting or task-specific reward models. 

Despite the benefits of parallel decoding, its effectiveness is constrained by substantial token consumption and the diminishing returns of majority voting: generating more traces does not always improve accuracy and may even degrade performance ~\cite{dynamic-voting, more-llm}. A key limitation is that traditional majority voting treats all reasoning traces equally, regardless of their quality ~\cite{hint-marginal, rank-voting}, so low-quality traces can disproportionately harm the final outcome.

Extending prior work, ~\citet{sets} unifies sampling, self-verification, and self-correction into a single test-time scaling pipeline, offering improved robustness over traditional approaches. However, these gains come at the cost of substantial token consumption and computational overhead, and the method still suffers from diminishing returns as the compute budget increases, which limits its scalability in practice. 
\begin{figure}[hbt]
  \centering
  \includegraphics[width=1.0\textwidth]{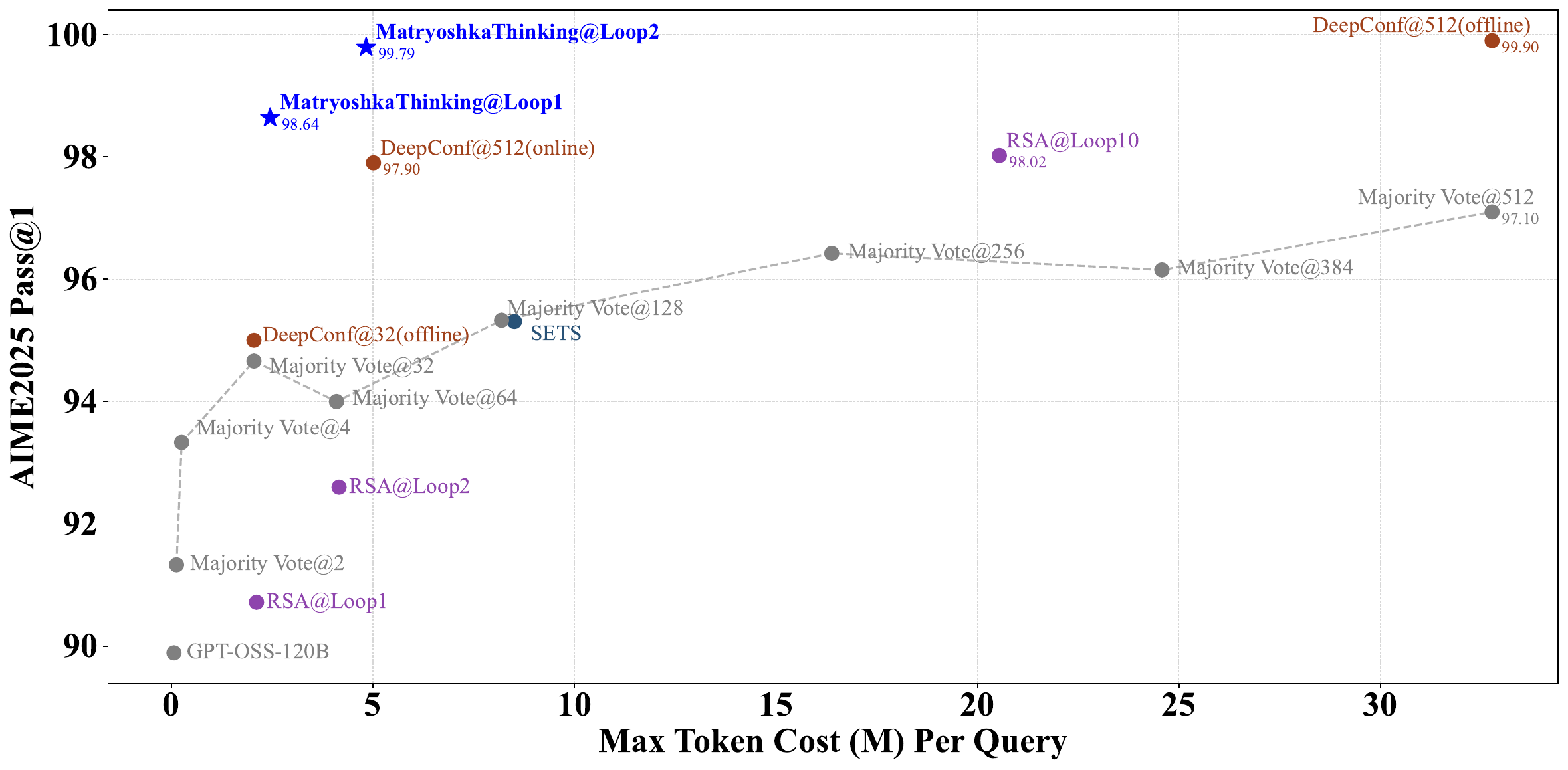}
  \caption{Performance–efficiency trade-off on AIME2025, showing pass@1 against inference token usage for different test-time scaling strategies base on GPT-OSS-120B. Efficient Reasoning via MatryoshkaThinking lies in the upper-left region, reflecting higher performance at lower inference cost. Each query is sampled 32 times.}
  \label{fig:teaser}
\end{figure}
In test-time scaling, recent research has concentrated on refining the strategies for selecting traces. ~\citet{b-o-n} offers a lightweight Best-of-N method that leverages token-level probabilities to enhance parallel decoding without reward models, though it remains vulnerable to overconfident errors and struggles on tasks with unique answers. Complementarily, DeepConf ~\cite{deepconf} introduces confidence-aware reasoning with dynamic self-reflection, yielding gains in multi-step reasoning but at the cost of higher computation and reliance on heuristic thresholding.

To address these challenges, we propose \textbf{MatryoshkaThinking}, an efficient and general reasoning enhancement method. It leverages the model’s inherent capabilities for sampling, judgment, and summarization to improve performance at inference time, without requiring additional training or auxiliary models, and can be seamlessly integrated into existing reasoning services. MatryoshkaThinking is applicable across different modalities of LLMs and demonstrates strong accuracy and efficiency on various multimodal benchmarks.
We evaluate MatryoshkaThinking on multiple reasoning benchmarks and models across different modalities. Experimental results show that MatryoshkaThinking significantly enhances reasoning performance while consuming fewer tokens compared to other test-time scaling methods.

\begin{figure}
  \centering
  \includegraphics[width=0.8\textwidth]{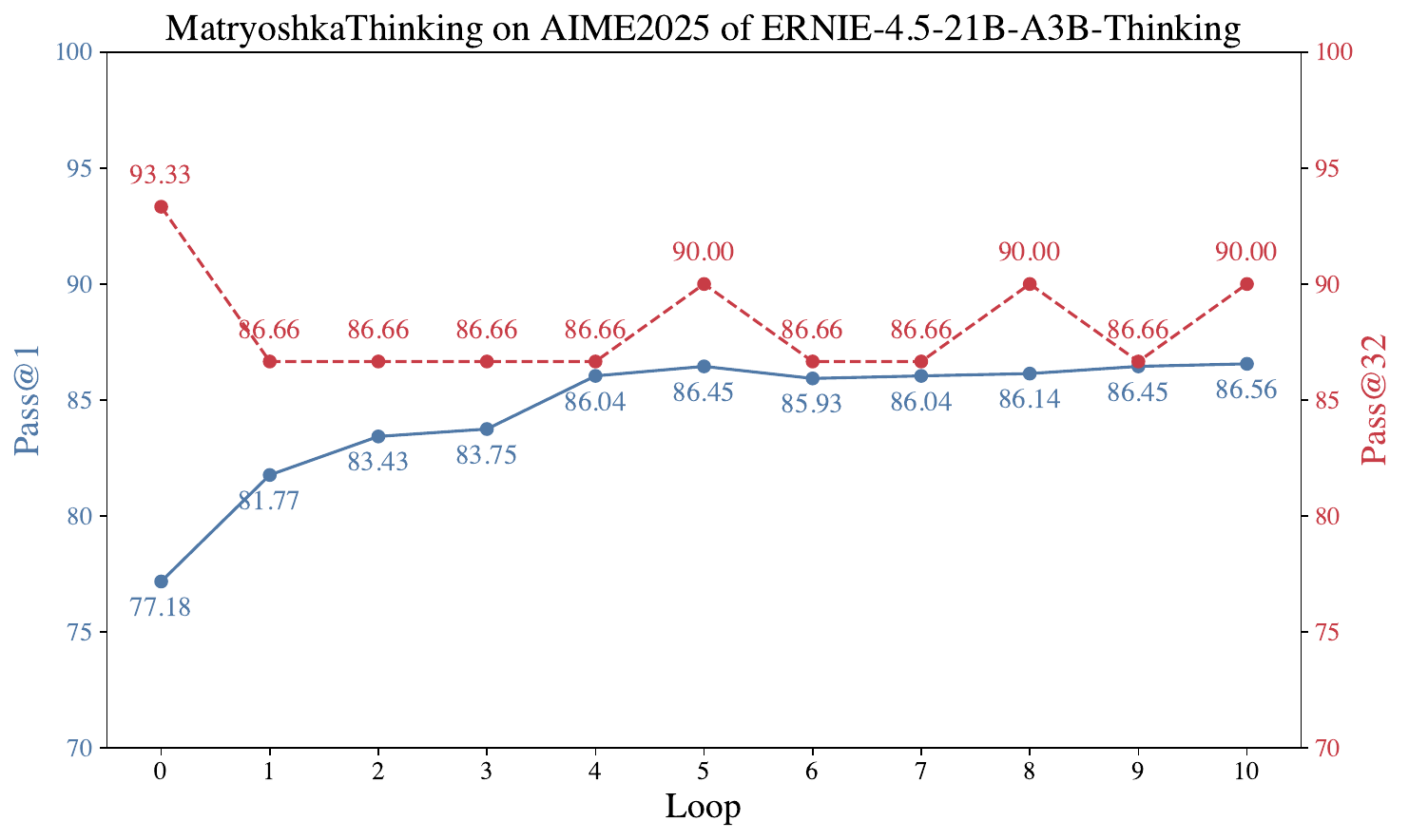}
  \caption{MatryoshkaThinking significantly enhances the Pass@1 accuracy of ERNIE-4.5-21B-A3B-Thinking on AIME2025, with performance converging toward Pass@32 as the number of recursive loop increases.
}
  \label{fig:pass@1scaling}
\end{figure}

In summary, our main contributions are as follows:
\begin{itemize}
    \item We propose a novel test-time scaling method, \textbf{MatryoshkaThinking}, which requires no additional models and relies solely on self-prompting to enhance model capabilities at inference. Under the same token budget, it outperforms existing methods such as DeepConf~\cite{deepconf} and SETS~\cite{sets}. Furthermore, our method demonstrates superior reasoning efficiency, as evidenced by the steeper performance scaling curve shown in Figure \ref{fig:teaser}. More importantly, MatryoshkaThinking effectively transfers the capability of Pass@k to Pass@1, enabling a single inference attempt to approach the performance level traditionally requiring 32 samples as Figure \ref{fig:pass@1scaling} shown.
    
    \item We conduct extensive experiments across multiple reasoning tasks and models in text, vision, and audio modalities. The results demonstrate that our method can significantly improve reasoning performance by increasing compute budget, highlighting its generality and broad applicability.
    
    \item We conducted extensive experiments, utilizing a total of \textbf{99.64B} tokens, which provided valuable insights into the challenges and potential of test-time scaling.
    
\end{itemize}

\section{Method}
\begin{figure*}[ht]
  \includegraphics[width=1.0\textwidth]{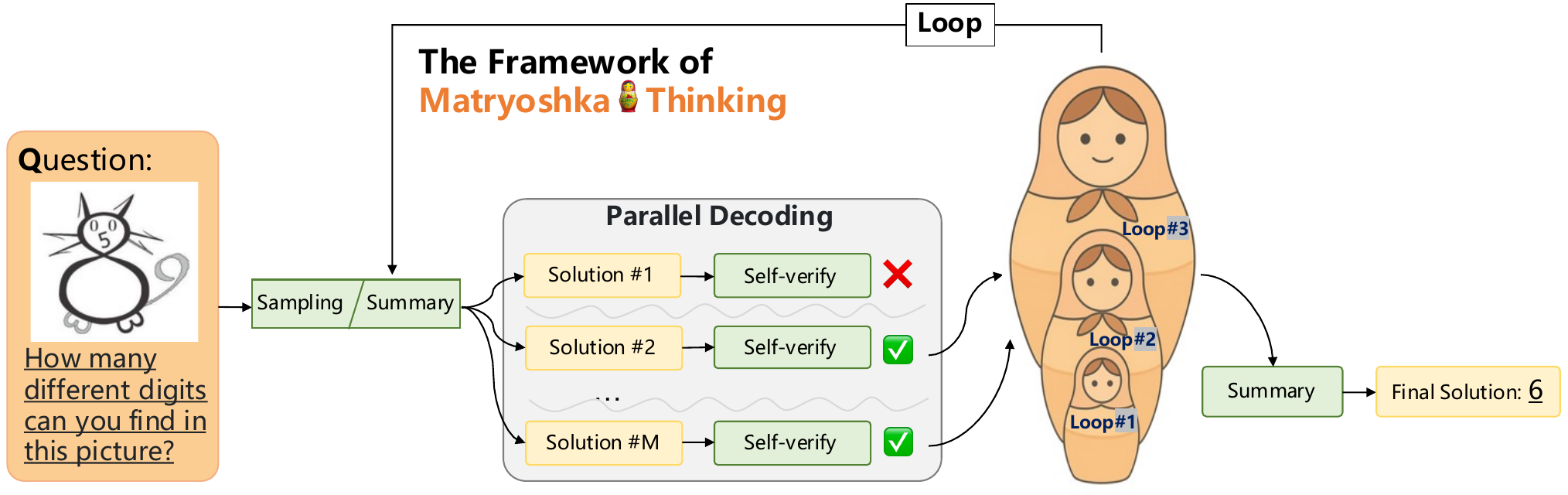}
  \caption{Overview of our MatryoshkaThinking. As shown in the figure, MatryoshkaThinking consists of three key components: self-verify, summary, and the iterative loop between them. With test-time scaling, recursive self-verify and summary identify promising solutions, ultimately consolidating them into a robust final solution.}
  \label{fig:overview}
\end{figure*}

We propose an inference method called \textbf{MatryoshkaThinking}, which leverages the intrinsic generative, discriminative, and summarization capabilities of large language models to achieve performance scaling through additional compute at test time. As illustrated in Figure~\ref{fig:overview}, MatryoshkaThinking begins with parallel sampling, in which multiple candidate responses are generated simultaneously for the same query. These candidates are then passed through a self-verify module that filters responses with model's internal consistency and logical validity. Based on the verification results, the model generates the summarization result that fuses reliable information from the retained candidates.

MatryoshkaThinking is extended with an iterative loop that repeats the summarize–verify cycle across multiple steps. In each iteration, the model first summarizes the accumulated set of previously verified outputs to generate a refined candidate set. This new set is then re-evaluated through the self-verification process. The accepted solutions are merged with prior results before the next summarization round. Through this iterative refinement and aggregation, the method progressively enhances solution quality and converges toward a high confidence final answer.
 
Specifically, we denote the answering prompt as \( I_a(x) \), the self-verify prompt as \( I_v(x, y) \), and the summary prompt as \( I_s\big(x, C \big) \). Here, \( x \) represents the input query, \(C\) represents solution candidates, \( y_l^m \) represents the \( m \)-th solution generated by the model in the \(l\)-th iteration for query \( x \), and \( v_l^m \) represents the model's self-verify result for the pair \( (x, y_l^m) \). As shown in Equation (\ref{eq:verification}), the judgment function \(J(\mathbf{v})\) outputs 1 if the solution is self-verified as correct, and 0 otherwise:

\begin{equation}
J(\mathbf{v}) =
\begin{cases} 
1 & \text{if } \mathbf{y} \text{ is self-verified as correct} \\
0 & \text{otherwise}
\end{cases}
\label{eq:verification}
\end{equation}
where $\mathbf{v}$ denotes the self-verify process.

Let \( \Theta \) represent the process of a large language model (LLM) taking a prompt as input and generating a response as output. In our MatryoshkaThinking framework, the iterative summary and verify procedures are formulated as follows:

\begin{align}
&y_0^m \sim \Theta(I_a(x)), \\
&y_l^m \sim \Theta(I_s(x, C_{l-1})), \\
&v_l^m \sim \Theta(I_v(x, y_{l}^m)), \\
&C_l = C_{l-1} \cup \{y_{l}^m \mid m = 1, \dots, M; J(v_l^m)=1\}
\end{align}

In this process, \(y_0^m\) is the initial candidate generated from the input \(x\). In each subsequent iteration \(l\), new candidates \(y_l^m\) are inferred based on the input \(x\) with the current candidate set \(C_{l-1}\) and then verified to obtain the self-verify result \(v_l^m\). The set \(C_l\) is updated with the candidates that have passed. The final answer is given by \(x\) with \(C_L\) produced in the final iteration \(L\). The pseudocode of our algorithm is summarized in Algorithm \ref{alg:matryoshka_thinking}.

\begin{algorithm}
\raggedright
\caption{MatryoshkaThinking Pipeline}
\label{alg:matryoshka_thinking}
\textbf{Input} The query $x$, the LLM \( \Theta \), the solution candidate set $C$, the answering prompt $I_a$, the self-verify prompt $I_v$, the summary prompt $I_s$, the number of loop $L$, the number of parallel decode samples $M$ and the judgement function $J$. 
\begin{algorithmic}[1]
\State $C \gets \emptyset$
\For{$l = 0$ to $L$}
    \For{$m = 1$ to $M$}   \Comment{\footnotesize Parallel Sampling}
        \If{$l = 0$}
            \State $y_l^m \leftarrow \Theta(I_a(x))$     \Comment{\footnotesize First Answer Step}
        \Else
            \State $y_l^m \leftarrow \Theta(I_s(x, C))$  \Comment{\footnotesize Summary Step}
        \EndIf
        \State $v_l^m \leftarrow \Theta(I_v(x, y_{l}^m))$ \Comment{\footnotesize Self-verify step}
    \EndFor
    \State $C \gets C \cup \{y_{l}^m|m = 1, \dots M; J(v_l^m)=1\}$  \Comment{\footnotesize Update Candidate Set}
\EndFor
\State $y_L \leftarrow \Theta(I_s(x, C))$
\end{algorithmic}
\textbf{Output} Final summarized answer $y_L$
\end{algorithm}

Departing from conventional voting-based aggregation strategies~\cite{sets, deepconf}, our method introduces a unified iterative framework that combines latent summarization and self-verification. This enables dynamic information integration and error correction across reasoning paths. The overall design provides three key advantages:

\begin{enumerate}[leftmargin=*,label=\textbf{\arabic*.}]
    \item \textbf{Broader Applicability}: A key advantage of the summarization approach lies in its flexibility. Unlike methods limited to tasks with enumerable or directly comparable outputs, such as classification and multiple-choice QA, it generalizes to open-ended problems with diverse and unconstrained answers. Majority voting, which relies on surface form agreement over a fixed output set, is ineffective in such contexts. In contrast, our method performs semantic-level aggregation by leveraging latent representations across multiple parallel reasoning paths, enabling robust inference across domains including generative question answering, creative text generation, and code generation.

    \item \textbf{Error Correction}: Another core advantage of the MatryoshkaThinking framework lies in its integration of self-verify ability, which serves as an internal filtering mechanism to improve answer quality. At each iteration, candidate outputs are subjected to self-verify before being retained, helping the model eliminate implausible or logically inconsistent solutions. This step mitigates error propagation across reasoning stages and ensures that only semantically consistent and logically outputs contribute to subsequent summarization. Unlike majority voting, which treats all candidates equally, self-verify introduces a quality control layer that significantly enhances the reliability and consistency of final predictions.

    \item \textbf{Robustness}: The combination of self-verification and latent summarization enables the model to remain robust even when no single candidate output is fully correct. Majority voting fails in such scenarios, as it merely amplifies the most common error. In contrast, our method integrates partial signals from multiple imperfect outputs at the representation level, allowing the model to recover coherent and high-quality solutions through iterative refinement.
\end{enumerate}

At its core, MatryoshkaThinking transforms traditional single-pass inference into an iterative, self-reflective process that unifies generation, self-verification, and summarization. By systematically filtering and refining candidate solutions across iterations, it enhances semantic coherence, correctness, and robustness. This design not only improves performance under fixed computational budgets, but also generalizes effectively to a wide range of reasoning-intensive tasks.

\section{Related Work}
\subsection{Large Reasoning Model}
Reasoning is a highly sought-after capability for large language models (LLMs). Early attempts to enhance reasoning abilities in LLMs primarily focused on mathematical problem-solving ~\citet{infer-with-math}, where outcome-based or process-based verifiers were trained to supervise the learning process ~\cite{stepbystep}. Chain-of-Thought (CoT) prompting ~\cite{cot} and its subsequent improvements ~\cite{self-consist, tot, got} have aimed to guide LLM in generating reasoning pathways that would lead to correct answers. More recent studies suggest that LLMs can emulate human-like System 1 (fast, intuitive) and System 2 (slow, deliberative) thinking ~\cite{system1}, enabling more effective handling of logical tasks. Reward-guided decoding approaches ~\cite{args, reward-guided, self-evaluation} have built on these foundational techniques by incorporating external or prospective feedback. The release of OpenAI-o1 ~\cite{gpto1} and DeepSeek-R1 ~\cite{deepseek-r1} has further fueled interest in test-time scaling techniques within the research community.

\subsection{Test-time Scaling}
Recent studies have shown that incorporating additional computational resources during inference can enhance the performance of large language models effectively ~\cite{meta-generation}. Test-time scaling methods can be broadly categorized into two approaches: parallel scaling and sequential scaling ~\cite{complex-tasks}.
Parallel scaling involves sampling multiple responses from the same model and aggregating them using operators such as self-consistency, adaptive-consistency, majority voting, reward model scoring, or confidence estimation ~\cite{self-consist,lm-monkey, deepconf, li2401escape, wan2024reasoning, wang2024make, wang2025sampling, wang2024soft, aggarwal2023let}. Recursive Self-Aggregation (RSA) improves reasoning ability by repeatedly aggregating multiple randomly sampled solutions generated by the model itself ~\cite{RSA}. Sequential scaling employs an iterative refinement strategy, using the model's own feedback to continuously improve responses until the output meets the predefined verification criteria ~\cite{self-refine,muennighoff2025s1}. Furthermore, when process-based verifier reward models are available, the test-time computation can be further expanded through strategies such as beam search or look-ahead search ~\cite{tts-optimally}. This work investigates efficient test-time scaling without relying on external reward models. ~\cite{muennighoff2025s1} leverages massive self-verification and self-improvement though extremely costly, achieving gold-medal performance in the International Mathematical Olympiad of 2025 and thus demonstrating the full potential of test-time scaling. 
We propose a simple yet effective approach that synergistically integrates parallel and sequential scaling strategies. This approach achieves superior performance compared to methods using either strategy alone. Although ~\citet{tts-optimally} also explored combining parallel sampling with sequential revision, their method requires training task-specific verifiers and revision models, which is often impractical due to the high cost of data collection. On the other hand, ~\cite{sets} also combines the two strategies, its self-correction mechanism involves multiple iterative loops, requiring substantial token consumption to achieve comparable accuracy, and its voting mechanism exhibits limitations in scenarios with insufficient answer diversity.
To validate the effectiveness of our approach, we conduct comprehensive and extensive evaluations. Unlike works such as ~\citet{tts-optimally} and ~\citet{deepconf}, which are evaluated solely on text-based benchmarks, our proposed MatryoshkaThinking method is systematically assessed on text, visual and audio benchmarks. Extensive experiments on various open-source models, including standard and thinking-enhanced models, demonstrate effectiveness and strong generalization capability of our approach.

\section{Experiment}
\subsection{Setup}
\subsubsection{\textbf{Benchmarks}}
To comprehensively evaluate the effectiveness of our method, we select a set of reasoning benchmarks across multiple modalities. 
For text reasoning tasks, we employ AIME 2024~\cite{aime}, AIME 2025~\cite{aime}, MMLU~\cite{mmlu}, and LiveCodeBench~\cite{livecodebench}, which collectively evaluate the model's competencies in logical reasoning, factual inference, and algorithmic implementation. For image reasoning tasks, we adopt MathVista~\cite{mathvista} and MMMU~\cite{mmmu}, both of which require cross-modal reasoning and complex problem-solving abilities. Finally, for audio reasoning tasks, we evaluate on MMSU~\cite{voicebench} and MMAU~\cite{mmau}, which assess the reasoning ability of audio modalities.

\subsubsection{\textbf{Baselines}}
To ensure the fairness of our comparison, we restrict our analysis to methods that have not undergone additional training or utilized external reward models. Specifically, we adopt \textbf{DeepConf}~\cite{deepconf}, \textbf{SETS}~\cite{sets} and ~\textbf{RSA}~\cite{RSA} as our main baselines. DeepConf enhances reasoning efficiency and performance by leveraging local confidence signals to dynamically filter low-quality reasoning traces. SETS, on the other hand, integrates parallel and sequential strategies by unifying sampling, self-verification, and self-correction into a single framework, thereby improving test-time reasoning. In SETS, we configured sampling 32 times and self-correction 3 rounds, while in RSA, we used a trajectory population of 32 and random sampled 2 candidate solutions, as adding more candidates without processed reasoning output would exceed the max sequence length. Three methods have significantly advanced test-time scaling and provide strong baselines for evaluating our approach. We summarize the key differences between our method and the baselines, as shown in Table \ref{tab:compare with baselines}.

\begin{table}
    \centering
  \caption{Comparison of different baselines with MatryoshkaThinking.}
  \label{tab:compare with baselines}
  \scalebox{1.0}{
    \begin{tabular}{ccccc}
    \toprule
    Method&Sampling&Self-Verify&Summay&Loop\\
    \midrule
    DeepConf ~\cite{deepconf} &{\color{green}\ding{51}}& {\color{green}\ding{51}} &{\color{red}\ding{55}} & {\color{red}\ding{55}}\\
    SETS ~\cite{sets} & {\color{green}\ding{51}} & {\color{green}\ding{51}} & {\color{red}\ding{55}} & {\color{red}\ding{55}}\\
    RSA ~\cite{RSA} & {\color{green}\ding{51}} & {\color{red}\ding{55}} & {\color{green}\ding{51}} & {\color{green}\ding{51}}\\
    MatryoshkaThinking & {\color{green}\ding{51}} & {\color{green}\ding{51}} & {\color{green}\ding{51}} & {\color{green}\ding{51}}\\
  \bottomrule
\end{tabular}
}
\end{table}

\subsubsection{\textbf{LLM Configs}}
For the multi-modal evaluation, we selected strong open-source models that represent the state of the art in their respective domains. In textual reasoning tasks, the thinking models include GPT-OSS-120B ~\cite{gpt-oss}, Seed-OSS-36B-Instruct ~\cite{seed-oss}, Qwen3-30B-A3B-Thinking-2507 ~\cite{qwen3}, Qwen3-4B-Thinking-2507 ~\cite{qwen3}, and ERNIE-4.5-21B-A3B-Thinking ~\cite{ernie4.5}. For comparison, we also considered non-thinking models, namely Qwen3-4B-Instruct-2507 ~\cite{qwen3} and ERNIE-4.5-21B-A3B ~\cite{ernie4.5}. For vision-language tasks, we used ERNIE-4.5-VL-28B-A3B ~\cite{ernie4.5}, while for audio-language tasks, we adopted Qwen-2.5-Omni-7B ~\cite{qwen-2.5-omni}. This setup ensures a systematic assessment of our approach across different modalities and both thinking and non-thinking LLMs.

\begin{table}
\centering
  \caption{Hyper-parameters of different models in our experiments.}
  \label{tab:hyper-parameters}
  \scalebox{1.0}{
    \begin{tabular}{ccccc}
    \hline
    \textbf{Model} & \textbf{Temperature} & \textbf{Top-p} & \textbf{Top-k} & \textbf{Max-tokens} \\
    \hline
    GPT-OSS-120B 
    & 1.0 & 1.0 & -1 & 64,000\\
    Seed-OSS-36B-Instruct
    & 1.1 & 0.95 & -1 & 64,000 \\
    Qwen3-30B-A3B-Thinking-2507
    & 0.6 & 0.95 & -1 & 64,000  \\
    Qwen3-30B-A3B-Instruct-2507
    & 0.6 & 0.95 & -1 & 64,000  \\
    Qwen3-4B-Thinking-2507
    & 0.7 & 0.8 & -1 & 64,000  \\
    Qwen3-4B-Instruct-2507
    & 0.7 & 0.8 & -1 & 64,000  \\
    ERNIE-4.5-21B-A3B-Thinking & 0.9 & 0.9 & -1 & 64,000 \\
    ERNIE-4.5-VL-28B-A3B & 0.2 & 0.8 & -1 & 64,000 \\
    Qwen2.5-Omni-7B & 0.9 & 1.0 & -1 & 8192 \\
    \hline
\end{tabular}
}
\end{table}

The hyper-parameters in our experiments of different models are shown in Table \ref{tab:hyper-parameters}.

\subsubsection{\textbf{Metric Settings}}
In our experiments, we evaluate each benchmark dataset using the pass@k~\cite{humaneval} metric as the equoation (\ref{pass@k}) shown:
\begin{equation}\label{pass@k}
\text{pass@}k := \mathbb{E}_{\text{Problems}} \left[ 1 - \frac{\binom{n-c}{k}}{\binom{n}{k}} \right]
\end{equation}
Here, $n$ denotes the total number of candidate solutions generated for a given problem, $c$ represents the number of correct solutions, $k$ is the number of sampled solutions, and $\binom{\cdot}{\cdot}$ refers to the binomial coefficient. Intuitively, pass@k measures the probability that at least one correct solution is included when randomly sampling $k$ solutions from $n$ candidates.
In addition to LiveCodeBench, we use the LLM-as-a-Judge~\cite{llm-as-judge} approach to evaluate the correctness of the answers.
\noindent\textbf{REMARK} Unless otherwise specified, all experimental results of \textsc{MatryoshkaThinking} correspond to the setting with \textit{Loop} = 2.
\subsection{Results}
We evaluated our method against several representative test-time scaling baselines, including DeepConf@32 (offline)~\cite{deepconf} (64M tokens, Pass@1 = 95.00), SETS~\cite{sets} (272M tokens, Pass@1 = 95.31), and RSA~\cite{RSA} (23M tokens, Pass@1 = 98.02). Each query is sampled 32 times. Voting-based methods must select a final answer under specific conditions for each query, which scales computation by a factor of 32 when producing 32 answers. In contrast, our method conducts 32 condition-based summarizations in an additive rather than multiplicative fashion, yielding a substantial reduction in computational cost.

As shown in Table \ref{tab:main result aime25}, our method achieves Pass@1 of 99.79 while consuming only 42M tokens.
Compared with SETS, it delivers a +4.48 absolute improvement in Pass@1 while using merely 15\% of its 272 M token cost.
In addition, our method attains near-perfect accuracy comparable to DeepConf@512 (offline) yet requires only 4\% of its computational budget.
Relative to RSA, our approach uses slightly more tokens (42 M vs. 23 M) but yields a +1.77 higher Pass@1, further highlighting its superior accuracy–efficiency trade-off.
The performance of RSA is restricted by the use of random sampled candidates rather than self-verify selection and by the unrefined processing of reasoning outputs. Self-verify enables our method to gather numerous robust candidate solutions, leading to improved summary quality.
These results demonstrate that our approach achieves near-optimal accuracy at a fraction of the cost, outperforming all prior methods in both efficiency and effectiveness. Moreover, it bridges the gap between low-cost approaches like RSA and high-accuracy methods like DeepConf@512 (offline), showing strong adaptability across computational budgets and highlighting its potential for large-scale deployment.

\begin{table}[htbp]
\centering
  \caption{Comparison of different test-time scaling methods on GPT-OSS-120B evaluated on the AIME2025 dataset. 
The table reports Pass@1 performance and the max token cost between different test-time scaling methods. The result of DeepConf@512 is taken from the original paper~\cite{deepconf}.}
  \label{tab:main result aime25}
  \scalebox{1.0}{
  \begin{tabular}{cccc}
    \toprule
    & \textbf{Method} & \textbf{Pass@1} & \textbf{Max\;Token\;Cost}\\
    \midrule
        & MajorityVote@32 & 94.66 & 64M \\ 
        & MajorityVote@128 & 95.33 & 262M \\  
        & MajorityVote@512 & 97.00 & 1024M \\
        & DeepConf@32 (offline) & 95.00 & 64M \\   
        & DeepConf@512 (online) & 97.90 & 160M \\   
        & DeepConf@512 (offline) & 99.90 & 1048M \\
        & SETS & 95.31 & 272M \\
        & RSA & 98.02 & 23M \\
        & MatryoshkaThinking & 99.79 & 42M \\
    \bottomrule
  \end{tabular}
  }
\end{table}

Next, we evaluate the generality of our approach in the AIME2025 dataset by applying MatryoshkaThinking to a variety of models. As shown in Table~\ref{tab:main result aime25 cross different model} all models achieve consistent improvements in Pass@1 over their baselines. Specifically, GPT-OSS-120B improves by +9.90, Seed-OSS-36B-Instruct by +3.23, Qwen3-30B-A3B-Thinking-2507 by +4.48, Qwen3-4B-Thinking-2507 by +7.94, and ERNIE-4.5-21B-A3B-Thinking by +9.38. Nonthinking models also benefit substantially as shown in Table \ref{tab:main result aime25 cross different model}, with Qwen3-30B-A3B-Instruct-2507 improving by +26.04 and Qwen3-4B-Instruct-2507 improving by +18.02. These results confirm that MatryoshkaThinking reliably boosts Pass@1 across diverse model families, demonstrating both its robustness and its broad applicability.

\begin{table}[htb]
\centering
  \caption{Performance of MatryoshkaThinking on AIME2025 across different models. * indicates that MatryoshkaThinking is set with loop=10; otherwise, the setting is loop=2.}
  \label{tab:main result aime25 cross different model}
  \scalebox{0.8}{
  \begin{tabular}{c|cccccc}
  \hline
    Type & Model & MatryoshkaThinking & Pass@1 & Pass@8 & Pass@16 & Pass@32 \\
    \hline
    \multirowcell{10}{Thinking Model} & 
        \multirow[c]{2}{*}{GPT-OSS-120B} & 
            {\color{red}\ding{55}} & 89.89 & 96.78 & 98.81 & 100.00 \\
        & & {\color{green}\ding{51}} & 99.79 & 100.00 & 100.00 & 100.00 \\
        \cline{2-7}
        & \multirow[c]{2}{*}{Seed-OSS-36B-Instruct} &
            {\color{red}\ding{55}} & 84.79 & 90.49 & 92.35 & 93.33 \\
        & & {\color{green}\ding{51}} & 88.02 & 92.62 & 93.26 & 93.33 \\
        \cline{2-7}
        & \multirow[c]{2}{*}{Qwen3-30B-A3B-Thinking-2507} &
            {\color{red}\ding{55}} & 84.58 & 92.83 & 93.30 & 93.33 \\
        & & {\color{green}\ding{51}} & 89.06 & 90.00 & 90.00 & 90.00 \\
        \cline{2-7}
        & \multirow[c]{2}{*}{Qwen3-4B-Thinking-2507} &
            {\color{red}\ding{55}} & 81.56 & 89.65 & 89.98 & 90.00 \\
        & & {\color{green}\ding{51}} & 89.50 & 90.00 & 90.00 & 90.00 \\
        \cline{2-7}
        & \multirow[c]{2}{*}{ERNIE-4.5-21B-A3B-Thinking\textsuperscript{*}} &
            {\color{red}\ding{55}} & 77.18 & 90.40 & 92.44 & 93.33 \\
        & & {\color{green}\ding{51}} & 86.56 & 87.50 & 88.33 & 90.00 \\
        \cline{2-7}
    \hline
    \multirowcell{4}{Non-Thinking Model} & 
        \multirow[c]{2}{*}{Qwen3-30B-A3B-Instruct-2507} &
            {\color{red}\ding{55}} & 43.64 & 60.14 & 64.98 & 73.33 \\
        & & {\color{green}\ding{51}} & 69.68 & 70.00 & 70.00 & 70.00 \\
        \cline{2-7}
        & \multirow[c]{2}{*}{Qwen3-4B-Instruct-2507} &
            {\color{red}\ding{55}} & 37.29 & 55.52 & 59.15 & 63.33 \\
        & & {\color{green}\ding{51}} & 55.31 & 59.97 & 60.00 & 60.00 \\
        \cline{2-7}
        \hline
\end{tabular}
  }

\end{table}

Furthermore, we extend MatryoshkaThinking to a broad spectrum of reasoning tasks spanning text-only, vision–language, and audio–language scenarios. As shown in Table~\ref{tab:main result on different modality},  on text-based reasoning with GPT-OSS-120B, our method improves Pass@1 by +0.96 on MMLU and +1.6 on LiveCodeBench. For vision–language reasoning, using ERNIE-4.5-VL-28B-A3B, we observe gains of +1.74 on MathVista and +2.55 on MMMU. For audio–language reasoning, with Qwen2.5-Omni-7B as the base model, Pass@1 increases by +7.57 on MMSU and +5.83 on MMAU. These consistent improvements across diverse modalities demonstrate the robustness and broad applicability of MatryoshkaThinking.

\begin{table}[htbp]
\centering
  \caption{MatryoshkaThinking consistently achieves gains across all modalities, with the largest improvement being +7.57 on MMSU, demonstrating the method’s effectiveness and generality. MMLU and LiveCodeBench V6 results are reported using GPT-OSS-120B. MathVista and MMMU results are reported using ERNIE-4.5-VL-28B-A3B. MMSU and MMAU results are reported using Qwen2.5-Omni-7B.}
  \label{tab:main result on different modality}
  \begin{tabular}{ccc}
  \hline
    \textbf{Benchmark} & \textbf{MatryoshkaThinking} & \textbf{Accuracy} \\
    \hline
    \multicolumn{3}{c}{\textit{\textbf{Text} Reasoning \& Coding Tasks (GPT-OSS-120B) }} \\
    \hline
        \multirow[c]{2}{*}{MMLU} & 
            {\color{red}\ding{55}} & 87.13 \\
        & {\color{green}\ding{51}} & 88.09 \\
        \multirow[c]{2}{*}{LiveCodeBench V6} &
            {\color{red}\ding{55}} & 88.46 \\
        & {\color{green}\ding{51}} & 90.06 \\
    \hline
    \multicolumn{3}{c}{\textit{\textbf{Vision} Reasoning Tasks (ERNIE-4.5-VL-28B-A3B)}} \\
    \hline
        \multirow[c]{2}{*}{MathVista} &
            {\color{red}\ding{55}} & 76.92 \\
        & {\color{green}\ding{51}} & 78.66 \\
        \multirow[c]{2}{*}{MMMU} &
            {\color{red}\ding{55}} & 68.55 \\
        & {\color{green}\ding{51}} & 71.10 \\        
    \hline
    \multicolumn{3}{c}{\textit{\textbf{Audio} Reasoning Tasks (Qwen2.5-Omni-7B)}} \\
    \hline
        \multirow[c]{2}{*}{MMSU} &
            {\color{red}\ding{55}} & 58.54 \\
        & {\color{green}\ding{51}} & 66.11  \\
        \multirow[c]{2}{*}{MMAU} &
            {\color{red}\ding{55}} & 70.08 \\
        & {\color{green}\ding{51}} & 75.91 \\
        \hline
\end{tabular}
\end{table}

\subsection{Ablation Study}
Our previous experiments have established that MatryoshkaThinking efficiently enhances reasoning capabilities by increasing the test-time computation, while consuming significantly fewer tokens than baseline methods. Building on these findings, we now turn to a deeper analysis of their implications for advancing test-time scaling.

\begin{table}[ht]
\centering

\caption{Ablation study of component contributions for GPT-OSS-120B on AIME2024 and AIME2025.}
\begin{tabular}{lcc}
\toprule
Method & AIME2024 (Pass@1) & AIME2025 (Pass@1)  \\
\midrule
MatryoshkaThinking & 99.89 & 99.79 \\
w/o Loop & 98.12 & 98.64 \\
w/o Loop \& Self-Verify & 97.18 & 96.45 \\
w/o Loop \& Self-Verify \& Summary. & 90.31 & 89.89 \\
\bottomrule
\end{tabular}

\label{tab:ablation_study}
\end{table}

\subsubsection{\textbf{Analyzing the Relative Contribution of Components}}
To quantify the contribution of each core component in MatryoshkaThinking, we conducted step-wise ablation experiments on the AIME2024 and AIME2025 benchmarks using the GPT-OSS-120B model.
Specifically, we progressively removed the Loop, Self-Verify, and Summary modules to examine their individual and cumulative impacts on performance.
As shown in Table~\ref{tab:ablation_study}, removing the Loop mechanism caused moderate degradation in Pass@1 accuracy—dropping from 99.89\% → 98.12\% (-1.77) on AIME2024 and 99.79\% → 98.64\% (-1.15) on AIME2025.
Further excluding the Self-Verify component resulted in a sharper decline to 97.18\% (-2.71) and 96.45\% (-3.34), respectively.
Finally, removing all three modules, including Summary, led to substantial drops to 90.31\% (-9.58) on AIME2024 and 89.89\% (-9.90) on AIME2025.
These results clearly show that each module contributes critically to the overall reasoning performance of MatryoshkaThinking.

\subsubsection{\textbf{Summary is not only Summary}}
\begin{figure}[htbp]
\centering
  \includegraphics[width=0.6\textwidth]{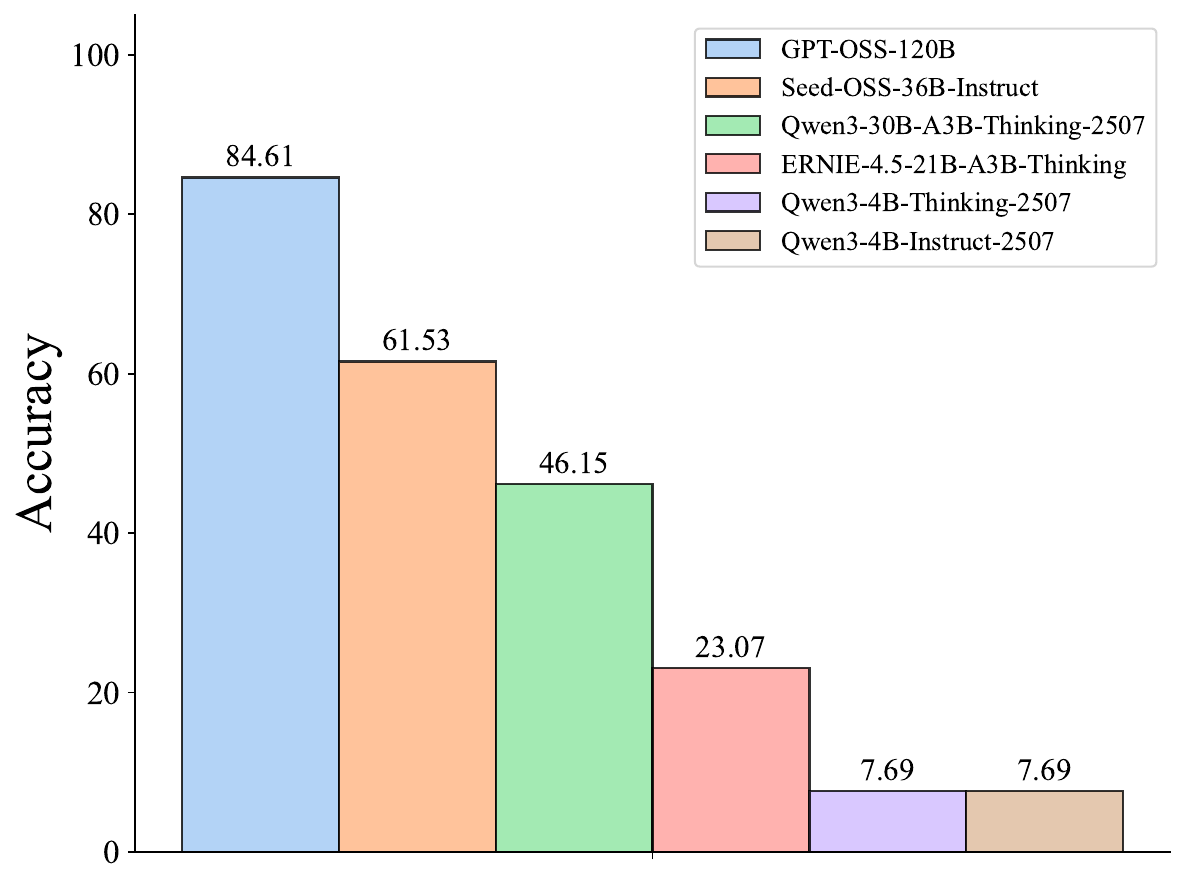}
  \caption{The figure illustrates the accuracy of various models when tasked with summarizing context without any correct answers. The results reveal that stronger models like GPT-OSS-120B (achieving 84.61\% accuracy) are able to correctly identify answers based on reasoning and error correction, while weaker models such as Qwen3-4B-Instruct-2507 (scoring 7.69\%) show limited or no inductive reasoning ability. These findings highlight the critical role of the Summary phase in enhancing model performance through inductive reasoning and error correction, especially for stronger models.}
  \label{fig:summary ability}
\end{figure}

In MatryoshkaThinking, the summary component not only compresses content but also provides inductive reasoning and error correction. 
We designed an experiment where we selected 24 questions and constructed Summary context with varying proportions of correct solutions to examine their effect on the model's accuracy performance.

As shown in Figure \ref{fig:summary ability}, when we supply only incorrect solution candidates and instruct the model to summarize them, models can still deduce the correct answer. For instance, GPT-OSS-120B achieves an accuracy of 84.61\% despite the summary context containing no correct answers. Other models also exhibit non-zero accuracy: Seed-OSS-36B-Instruct attains 61.53\%, Qwen3-30B-A3B-Thinking-2507 reaches 46.15\%, ERNIE-4.5-21B-A3B-Thinking scores 23.07\%, while both Qwen3-4B-Thinking-2507 and Qwen3-4B-Instruct-2507 achieve 7.69\%. These results demonstrate that during the summary phase, models engage not merely in summarization, but also in inductive reasoning and error correction. Furthermore, these additional capabilities vary significantly depending on model capability.

\subsubsection{\textbf{Investigating the Impact of Varying Proportions of Correct Answers in Context}}

\begin{figure}[htbp]
\centering
  \includegraphics[width=0.6\textwidth]{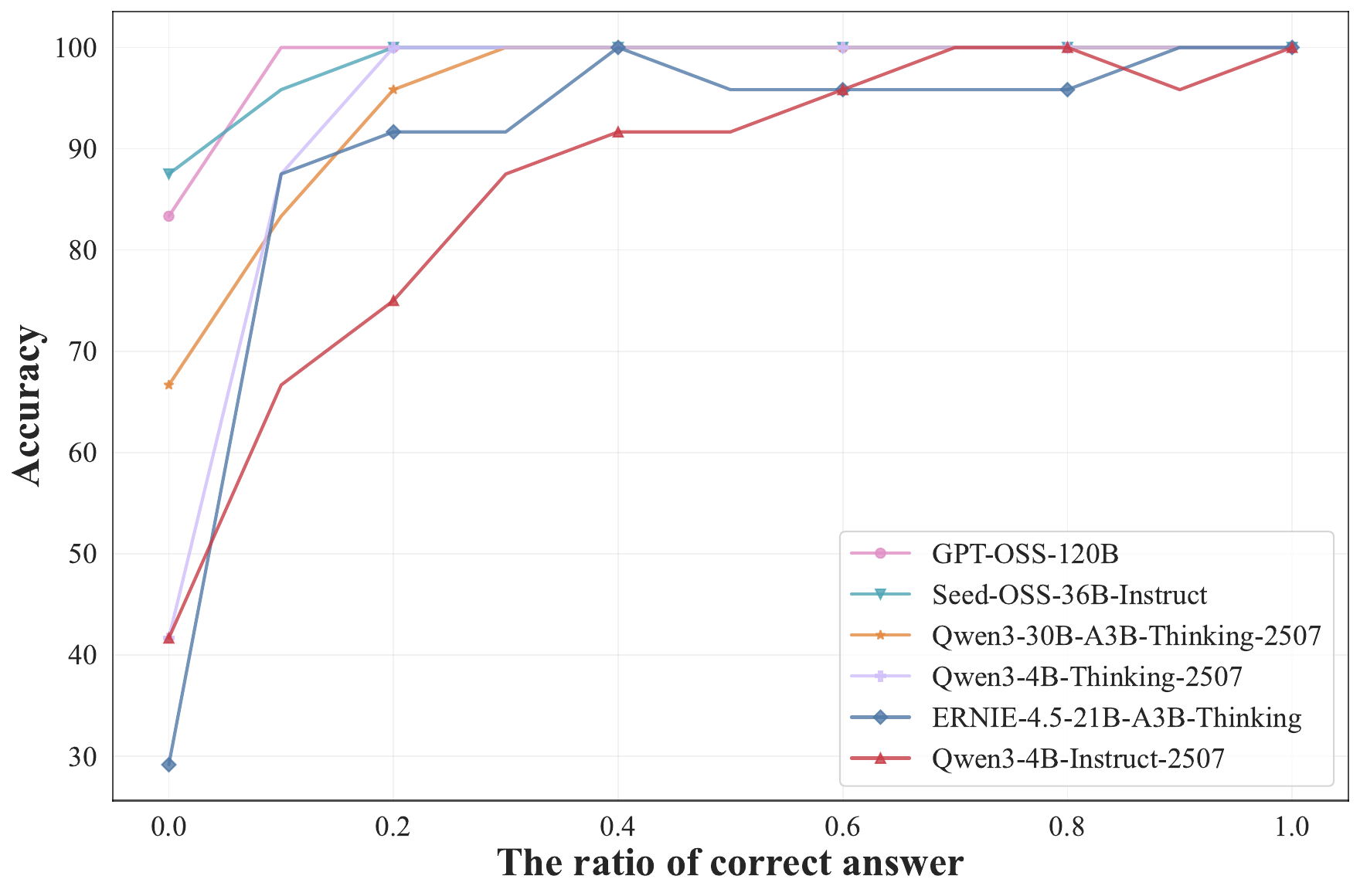}
  \caption{The figure demonstrates that, as the proportion of correct answers in the summary context increases, most models performance improves. In particular, for stronger models like GPT-OSS-120B, the summary method outperforms majority voting, especially when the proportion of correct answers in the context is small. This ratio of correct answers simulates the model’s self-verify capability in real-world applications, highlighting that the model's verification ability is crucial for improving performance.}
  \label{fig:verify_ability}
\end{figure}

\begin{figure}[htbp]
    \centering
    \begin{subfigure}[b]{0.43\textwidth}
        \centering
        \includegraphics[width=\textwidth]{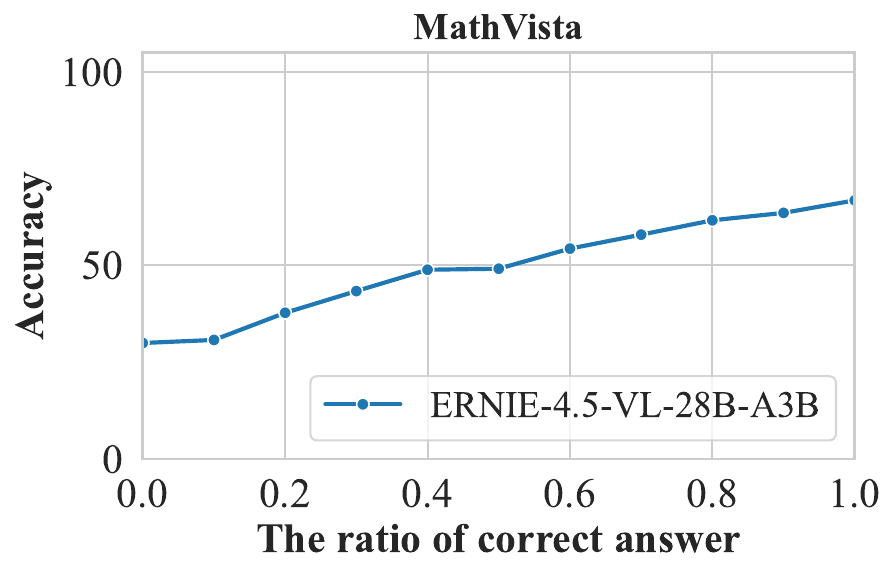}
        \caption{}
        \label{subfig:vl_mathvista}
    \end{subfigure}
    \hfill
    \begin{subfigure}[b]{0.43\textwidth}
        \centering
        \includegraphics[width=\textwidth]{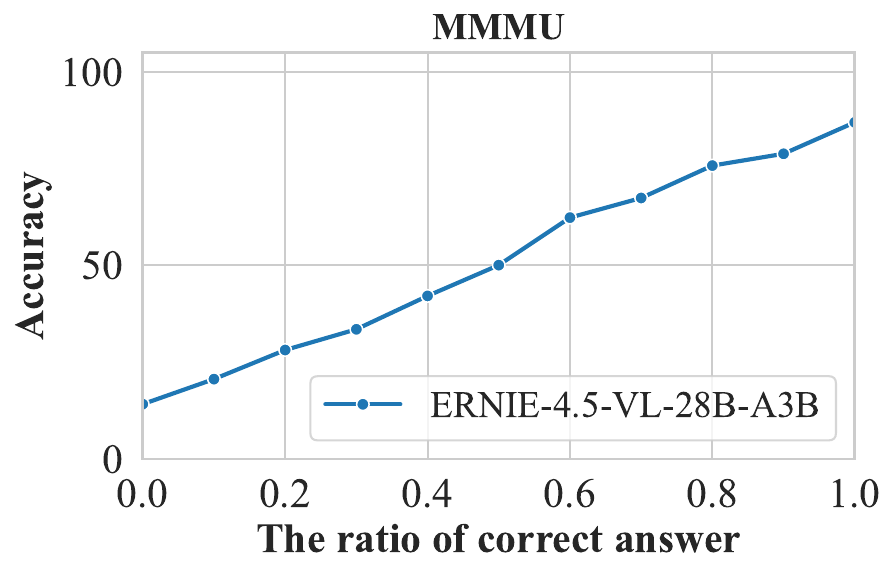}
        \caption{}
        \label{subfig:vl_mmmu}
    \end{subfigure}
    \hfill
    \begin{subfigure}[b]{0.43\textwidth}
        \centering
        \includegraphics[width=\textwidth]{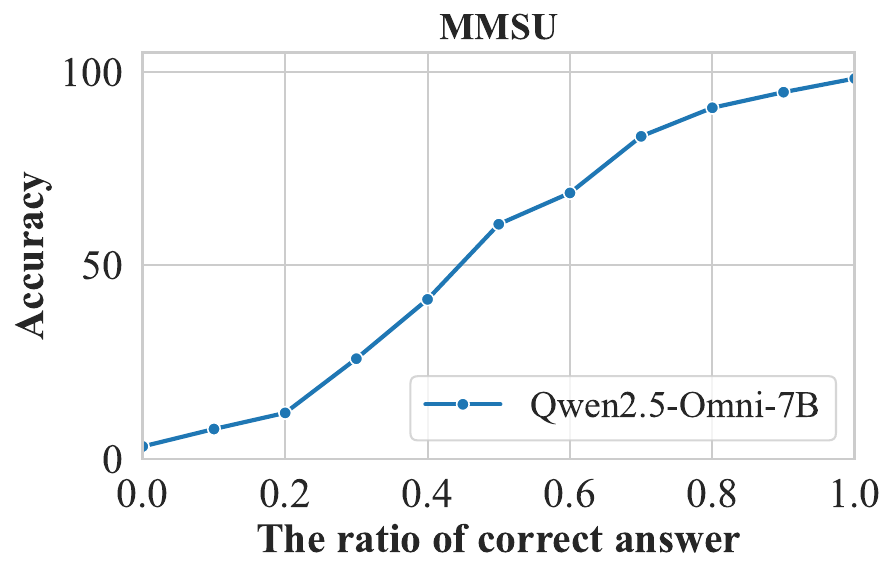}
        \caption{}
        \label{subfig:qwen_mmsu}
    \end{subfigure}
    \hfill
    \begin{subfigure}[b]{0.43\textwidth}
        \centering
        \includegraphics[width=\textwidth]{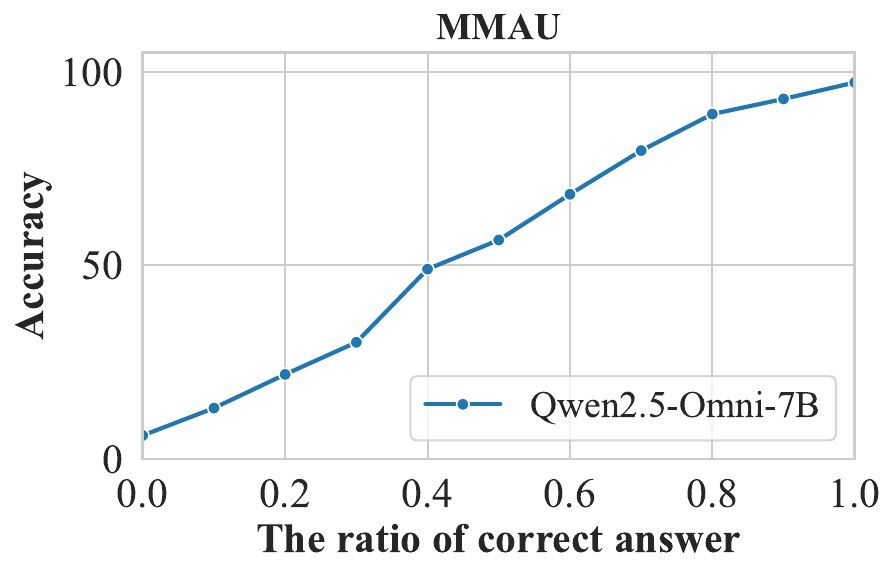}
        \caption{}
        \label{subfig:qwen_mmau}
    \end{subfigure}
    
    \caption{
        Scaling effects of verification quality in summarization across multimodal benchmarks. 
        (a) ERNIE-4.5-VL-28B-A3B on MathVista benchmark; 
        (b) ERNIE-4.5-VL-28B-A3B on MMMU benchmark; 
        (c) Qwen2.5-Omni-7B on MMSU benchmark; 
        (d) Qwen2.5-Omni-7B on MMAU benchmark. 
        Our findings indicate that higher correct solution ratios consistently lead to better Pass@1 performance, underscoring the critical role of verification quality across modalities.
    }
    \label{fig:verify_multimodal}
\end{figure}

As observed in earlier experiments, models achieved non-zero accuracy even in the absence of correct answers provided in the summary context. To further investigate this capability, we examined how model performance is influenced by varying the proportion of correct answers included in the summary context.

We designed an experiment where we selected 24 questions and constructed summary context with varying proportions of correct solutions to examine their effect on the model's accuracy performance. The results, as shown in Figure \ref{fig:verify_ability}, indicate that for strong models like GPT-OSS-120B, introducing just 10\% correct solutions into the Summary condition boosted the Pass@1 score to 100\%. In contrast, using Majority Voting for aggregation failed to achieve this level of performance. This shows that for stronger models, the summary method outperforms the Voting method, especially when aggregating multiple reasoning paths, as it better captures valuable answers.
However, for weaker models, such as Qwen3-4B-Instruct-2507, even with a 60\% correct answer ratio in the summary context, they still could not reach full accuracy. This suggests that, in the case of weaker models, the summary method does not always outperform voting. Nevertheless, we observed that in most cases, especially with stronger models, the summary method provided significantly better performance than voting.

The experimental results show that the ratio of correct solutions in the summary context reflects the self-verification capability of models in the practical application of MatryoshkaThinking. As illustrated in the Figure \ref{fig:verify_ability}, when the summary context contains a higher proportion of correct solutions, models with strong summarization abilities achieve greater improvements, while models with weaker summarization abilities also benefit but to a smaller extent. This indicates that the extent of improvement depends not only on self-verification but also on the model’s intrinsic summarization ability. In other words, self-verification acts as an enhancement rather than a remedy: it strengthens existing summarization capacity but cannot compensate for its absence.

To further examine whether the benefit of increasing the proportion of correct solutions in the summary context generalizes beyond textual reasoning, we conducted analogous experiments on visual and audio tasks. As shown in Figure~\ref{fig:verify_multimodal}. Using ERNIE-4.5-VL-28B-A3B for visual tasks and Qwen2.5-Omni-7B for audio tasks, we evaluated the models on benchmarks like MathVista, MMMU, MMSU, and MMAU. The results consistently showed that introducing just 10\% correct solutions into the Summary condition improved the performance across all benchmarks: MathVista (+0.79), MMMU (+6.46), MMSU (+4.50), and MMAU (+7.07). Moreover, further increasing the proportion of correct solutions continued to boost performance, underlining the crucial role of self-verification in enhancing MatryoshkaThinking's effectiveness, even in multimodal contexts.

\subsubsection{\textbf{More Loops Bring Stronger Confidence To Truth}}

\begin{figure}[hbt]
    \centering
    \includegraphics[width=0.6\textwidth]{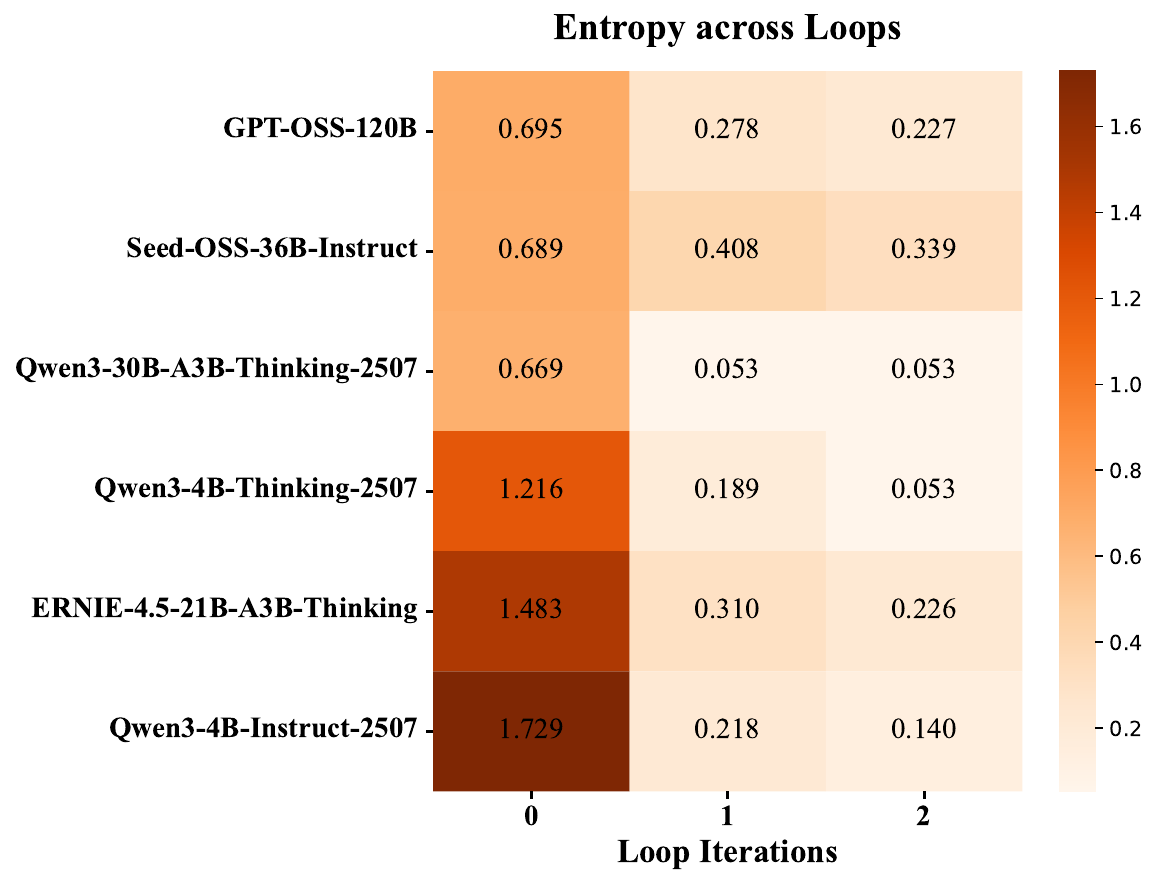}
    \caption{The color intensity represents the magnitude of entropy. As the number of loop iterations increases, the entropy of the model’s answers gradually decreases. This indicates that the model becomes more confident in its predictions, but at the same time, its ability to explore diverse solution paths is suppressed.}
    \label{fig:entropy_hotmap}
\end{figure}
To further investigate the impact of MatryoshkaThinking iterations on the uncertainty of model predictions, we conducted experiments on the AIME2025 dataset using GPT-OSS-120B. In this setup, the iteration variable is denoted as $Loop_N$, where $N \in [0, 2]$. At each iteration step, we calculated the entropy of the distribution about the candidate answers to quantify the model’s uncertainty in its predictions. The results, as shown in Figure~\ref{fig:entropy_hotmap}, reveal a clear trend: as $N$ increased from 0 to 2, the entropy of the generated answers decreased progressively. This indicates that additional iterations in MatryoshkaThinking effectively reduce uncertainty, thereby enhancing the model’s confidence in its predictions.

The reduction in entropy explains the observed improvement in Pass@1, as the model becomes more confident about its answers within each iteration. Furthermore, we also notice that when the number of iterations reaches 2, the entropy level plateaus, suggesting that further increasing the compute budget would yield diminishing returns. This trend helps explain why MatryoshkaThinking leads to an increase in Pass@1 for some models, while causing a decrease in Pass@32, as shown in Table \ref{tab:main result aime25 cross different model}. As entropy decreases and the model's confidence grows, its exploration capability is suppressed. Consequently, the model becomes less likely to generate diverse, potentially correct solutions across multiple reasoning traces, which leads to a drop in Pass@32.

\begin{figure}[hbp]
\centering
  \includegraphics[width=0.8\textwidth]{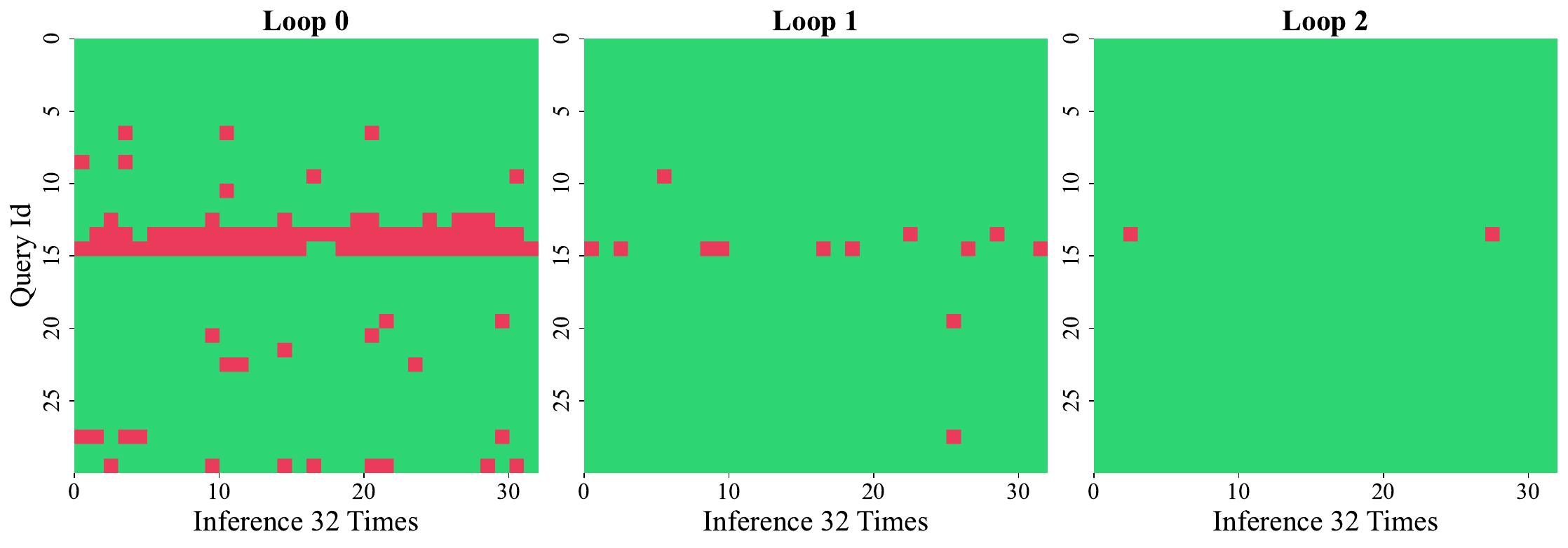}
  \caption{The evolution of the distribution of correct (green) and incorrect (red) answers from GPT-OSS-120B on AIME2025 with our MatryoshkaThinking across progressive loops.}
  \label{fig:loop accuracy}
\end{figure}

We further investigate the loop mechanism employed in MatryoshkaThinking. Using GPT-OSS-120B on the AIME2025 dataset as a case study, we track the distribution of correctly and incorrectly solved problems across successive loop iterations. As shown in Figure~\ref{fig:loop accuracy}, green cells denote questions that the model correctly answered in a certain loop, while red cells indicate incorrect responses. The model’s initial performance in loop 0 reflects its direct response capability.
It was particularly vulnerable to difficult questions — the accuracy on the 13th and 14th problems fell below 10\%. However, after a single iteration loop, performance improved significantly as the model began to take advantage of its previous solution history. This resulted in accuracy for these specific questions exceeding 50\%. In the second loop, the model solved almost all questions. The process illustrated in Figure~\ref{fig:loop accuracy} clearly demonstrates how the model’s predictions evolve across loop iterations. By recursively exploiting the model’s inherent strengths in reasoning, verification, and summarization, it progressively corrects earlier mistakes and recovers accurate solutions, even when valid answers constitute only a minority. This dynamic behavior further confirms the effectiveness of our approach. Building on this analysis, we attempt to model the relationship between MatryoshkaThinking’s accuracy and three key factors in the Appendix~\ref{app:formula}.

\section{Conclusion}
We present \textbf{MatryoshkaThinking}, a novel test-time scaling paradigm that operates solely on a model's intrinsic abilities without any fine-tuning or external feedback. 
The key advantage of our method lies in its efficiency and exceptional generality as evidenced by extensive evaluations on diverse models (from non-reasoning-model to reasoning-model) and tasks (from textual to multimodal understanding). 
Beyond presenting a practical framework, we conducted a comprehensive ablation study, revealing that improvements in iterative looping and summarization capabilities are pivotal for achieving effective test-time scaling. We hope that MatryoshkaThinking will serve as a foundational step towards more efficient and scalable inference techniques, inspiring future exploration in the community.

\section{Future Work}
While MatryoshkaThinking demonstrates broad generality by relying only on intrinsic model abilities, this design also limits self-verification in complex scenarios. A promising future direction is to integrate the framework with external tools such as web search, Python, or other APIs, potentially improving the reliability and scalability of test-time inference. Another potential avenue for future research is to integrate MatryoshkaThinking's parallel decoding capability into the model's intrinsic abilities, thereby enhancing both intelligence and reasoning efficiency.

\bibliography{colm2024_conference.bbl}
\bibliographystyle{colm2024_conference}

\clearpage
\appendix
\appendix
\section{Modeling the Interplay of Key Factors in MatryoshkaThinking}
\label{app:formula}

To better understand the effect of MatryoshkaThinking, we try to model the relation between the resulting accuracy of the method and three key factors: the model’s initial accuracy, verification capability, and summarization capability.

\subsection{Definitions}

\begin{definition}[Probability of Correct Solution]
The probability of generating a correct solution is 
\[
pass@1.
\]
\end{definition}

\begin{definition}[Verification Process]
In the verification process, given that a candidate solution may be either correct or incorrect, the verifier may classify it as correct with certain probabilities:
{\small
\[
\text{True Positive Rate (TPR)} = Pr[\text{verified as correct} \mid \text{solution is correct}],
\]
\[
\text{False Positive Rate (FPR)} = Pr[\text{verified as correct} \mid \text{solution is incorrect}].
\]
}
\end{definition}

\begin{definition}[Summarization Process]
In the summarization process, assume that given a set of candidate solutions, the probability of summarizing a correct one can be expressed as a function
\[
S(r),
\]
where \(r\) denotes the proportion of correct answers among the candidates.
\end{definition}

\subsection{Step1: Accuracy After Verification}

Suppose we randomly generate a solution, the probability that it is verified as the correct solution is:
\begin{equation}
Pr[\text{verified as correct}] = pass@1 \cdot TPR + (1 - pass@1) \cdot FPR
\end{equation}

Given that a solution is verified as correct, the probability that it is actually correct is:  
\begin{equation}
\begin{split}
q &= Pr[\text{solution is correct} \mid \text{verified as correct}] \\
  &= \frac{pass@1 \cdot TPR}{pass@1 \cdot TPR + (1 - pass@1) \cdot FPR}
\end{split}
\end{equation}

If there are $n_v$ candidate solutions verified as correct, the probability that the candidate set contains at least one correct solution is:  
\begin{equation}
P_{\text{has}} = 1 - (1 - q)^{n_v}
\end{equation}

Correspondingly, the probability that all candidates in the set are incorrect is:  
\begin{equation}
P_{\text{none}} = (1 - q)^{n_v}
\end{equation}

\subsection{Step2: pass@1 After Summarization}

Let us define:
\begin{itemize}
    \item $S(r = 0)$: the probability that summarization still produces a correct solution when the candidate set contains no correct solutions.
    \item $S(r > 0)$: the probability that summarization produce a correct solution when the candidate set contains at least one correct solution.
\end{itemize}

Therefore, with $n_v$ candidate solutions verified as correct, the $pass@1_{\text{final}}$ after summarization can be expressed as the sum of two parts:
\[
\begin{split}
pass@1_{\text{final}} &= P_{\text{has}} \cdot S(r > 0) + P_{\text{none}} \cdot S(r = 0) \\
&= \big(1 - (1 - q)^{n_v}\big) \cdot S(r > 0) 
   + (1 - q)^{n_v} \cdot S(n_v, r = 0) \\
&= \left(1 - \left(1 - \frac{p \cdot TPR}{p \cdot TPR + (1 - p) \cdot FPR}\right)^{n_v}\right) 
   \cdot S(r > 0) \\
&\quad + \left(\left(1 - \frac{p \cdot TPR}{p \cdot TPR + (1 - p) \cdot FPR}\right)^{n_v}\right) 
   \cdot S(r = 0)
\end{split}
\]
where $p = pass@1_{\text{init}}$.

\subsection{Step3: Binomial Distribution of Verified Candidate Set Size}

\textbf{Definition of the Binomial Distribution.} 
Let $X$ denote the number of successes in $n$ independent Bernoulli trials, each with success probability $p$. Then $X$ follows a binomial distribution with probability mass function:
\[
P(X = k) = \binom{n}{k} p^k (1-p)^{n-k}, \quad k = 0,1,\ldots,n
\]

Therefore, the number of candidate solutions after verification, denoted $N_v$, follows a binomial distribution.  
Suppose that initially one question generates $N$ solutions. Then the distribution of $N_v$ is:
{\small
\[
\begin{split}
P(N_v = n_v) = \\
&\binom{N}{n_v} Pr[\text{verified as correct}]^{n_v} \\
&\left(1 - Pr[\text{verified as correct}]\right)^{N-n_v}, \\
&\quad n_v = 0,1,\ldots,N
\end{split}
\]
}

Hence, the final expression for $pass@1_{\text{final}}$ is:
{\small
\[
\begin{split}
pass@1_{\text{final}} = \\
& \sum_{n_v=0}^{N} P(N_v = n_v) \Bigg[ \\
& \bigg(1 - \left(1 - 
\frac{pass@1_{\text{init}} \cdot TPR}{pass@1_{\text{init}} \cdot TPR + (1 - pass@1_{\text{init}})\cdot FPR}\right)^{n_v}\bigg) \\
& \cdot S(r > 0) \qquad \\
& + \left(1 - 
\frac{pass@1_{\text{init}} \cdot TPR}{pass@1_{\text{init}} \cdot TPR + (1 - pass@1_{\text{init}})\cdot FPR}
\right)^{n_v} \\
& \cdot S(r = 0) \\
& \Bigg]
\end{split}
\]
}

\end{document}